\pdfoutput=1

\documentclass[11pt]{article}

\usepackage[]{acl}

\usepackage{times}
\usepackage{latexsym}

\usepackage[T1]{fontenc}

\usepackage[utf8]{inputenc}

\usepackage{microtype}

\usepackage{xcolor}
\usepackage{graphicx}

\usepackage{booktabs}
\usepackage{multirow}
\usepackage{hhline}
\usepackage{subcaption}

\usepackage{tabularx}
\usepackage{arydshln}
\usepackage{pifont}
\usepackage{amsmath}

\title{Mitigating Catastrophic Forgetting in Large Language Models \\ 
with Self-Synthesized Rehearsal}

\author{Jianheng Huang\textsuperscript{1,3,5}\quad
    Leyang Cui\textsuperscript{2}\quad
    Ante Wang\textsuperscript{1,3,5}\quad
    Chengyi Yang\textsuperscript{1}\\
    \textbf{Xinting Liao\textsuperscript{4}\quad
    Linfeng Song\textsuperscript{2}\quad
    Junfeng Yao\textsuperscript{5}\quad
    Jinsong Su\textsuperscript{1,3,5}\thanks{~~Corresponding author.}} \\
    \textsuperscript{1}School of Informatics, Xiamen University\quad
    \textsuperscript{2}Tencent AI Lab\quad\\
    \textsuperscript{3}Shanghai Artificial Intelligence Laboratory\quad
    \textsuperscript{4}Zhejiang University\\
    \textsuperscript{5}Key Laboratory of Digital Protection and Intelligent Processing of Intangible Cultural Heritage \\
    of Fujian and Taiwan (Xiamen University), Ministry of Culture and Tourism, China\\
    \texttt{enatsu@stu.xmu.edu.cn}~~
    \texttt{jssu@xmu.edu.cn}
}
\begin{document}
\maketitle
\begin{abstract}
Large language models (LLMs) suffer from catastrophic forgetting during continual learning. Conventional rehearsal-based methods rely on previous training data to retain the model's ability, which may not be feasible in real-world applications. When conducting continual learning based on a publicly-released LLM checkpoint, the availability of the original training data may be non-existent. To address this challenge, we propose a framework called Self-Synthesized Rehearsal (SSR) that uses the LLM to generate synthetic instances for rehearsal. Concretely, we first employ the base LLM for in-context learning to generate synthetic instances. Subsequently, we utilize the latest LLM to refine the instance outputs based on the synthetic inputs, preserving its acquired ability. Finally, we select diverse high-quality synthetic instances for rehearsal in future stages. Experimental results demonstrate that SSR achieves superior or comparable performance compared to conventional rehearsal-based approaches while being more data-efficient. Besides, SSR effectively preserves the generalization capabilities of LLMs in general domains.
\end{abstract}

\section{Introduction}

Large language models (LLMs) have demonstrated remarkable performance across various natural language processing (NLP) tasks \cite{touvron2023llama2,openai2023gpt4}. 
In real-world applications, LLMs are often updated in a continual learning (CL) manner \cite{masson2019episodic}, where new instruction tuning data is incrementally introduced over time.
However, a significant issue that limits the effectiveness of LLMs is catastrophic forgetting, which refers to the LLM's tendency to forget previously acquired knowledge when learning new instances \cite{Kirkpatrick_2017,li-etal-2022-overcoming,luo2023empirical}.

To mitigate catastrophic forgetting, a line of work focuses on rehearsing previous training instances  \cite{masson2019episodic,rolnick2019experience,scialom-etal-2022-fine}.
These rehearsal-based methods maintain the model's ability by training on real data from previous training stages.
However, the real data may not always be desirable in practical applications. 
For instance, when conducting continual learning based on a publicly-released LLM checkpoint (e.g. Llama-2-chat), the availability of the original training data may be non-existent.
This raises an interesting research question: \textit{Can we maintain the LLM's ability during continual learning without using real data in previous training stages?}

\begin{figure}[tbp]
    \centering
    \includegraphics[width=\linewidth]{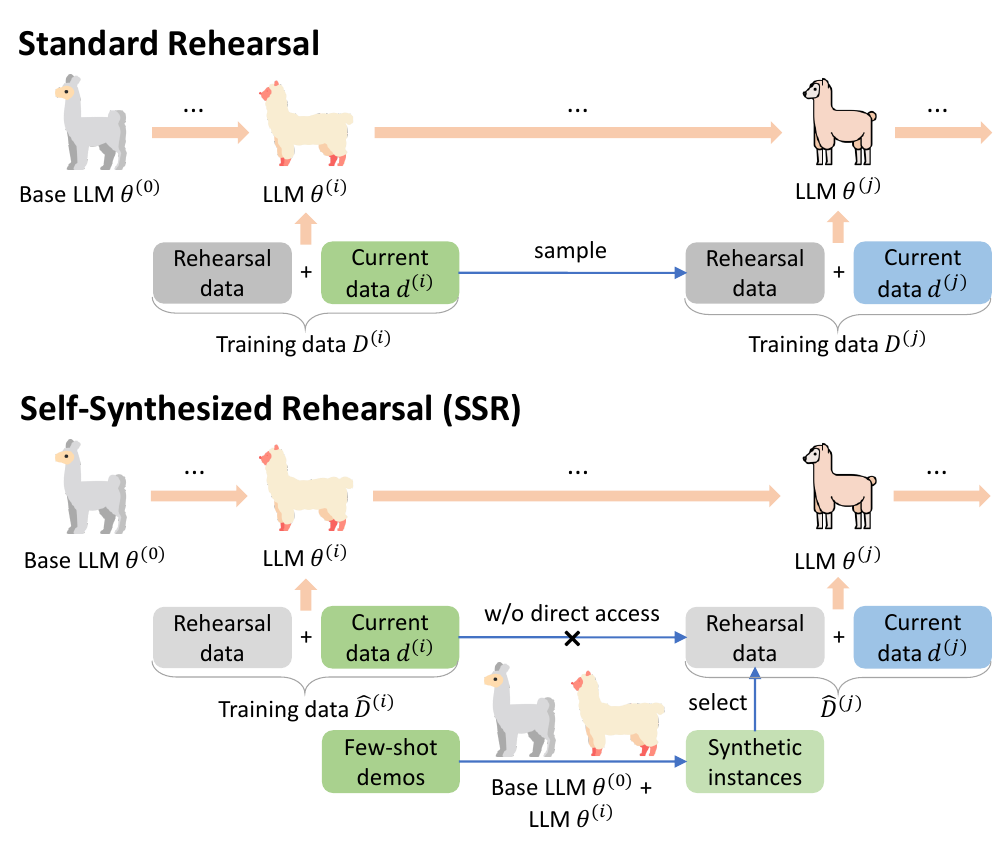}
    \caption{Comparison of standard rehearsal and our proposed Self-Synthesized Rehearsal (SSR).}
    \label{fig:diff}
\end{figure}

We propose the \textbf{S}elf-\textbf{S}ynthesized \textbf{R}ehearsal (\textbf{SSR}) framework to mitigate catastrophic forgetting in continual learning.
As shown in Figure~\ref{fig:diff}, unlike standard rehearsal-based continual learning that samples training instances from previous stages as rehearsal data, 
SSR framework uses the LLM to generate synthetic instances for rehearsal.
Specifically, we first use the base LLM to generate synthetic instances, conducting in-context learning (ICL) with few-shot demonstrations. These demonstrations can be collected from the previous data or human-constructed containing similar knowledge to the previous data. Then, the latest LLM is used to refine the outputs of synthetic instances to retain the latest LLM's ability. Finally, we select diverse high-quality synthetic instances for rehearsal in the future stages.

Extensive experiments on the task sequences derived from the SuperNI dataset \cite{wang-etal-2022-super} demonstrate that SSR has superior or comparable performance compared to the conventional rehearsal-based approaches, with higher data utilization efficiency.
Besides, experiments on AlpacaEval and MMLU \cite{hendrycks2021measuring} show that SSR can also effectively preserve the generalization capabilities of LLMs in general domains. 
We release our code and data at \url{https://github.com/DeepLearnXMU/SSR}.

\section{Related Work}

Learning a sequence of datasets continually while preserving past knowledge and skills is a crucial aspect of achieving human-level intelligence. Existing approaches to continual learning can be broadly categorized into three main categories: (i) regularization-based, (ii) architecture-based, and (iii) rehearsal-based methods. Regularization techniques \cite{Kirkpatrick_2017,cha2021cpr,huang2021continual,zhang-etal-2022-clle} control the extent of parameter updates during the learning process, preventing interference with previously learned tasks. Nonetheless, these methods typically rely on hyperparameters that need to be carefully tuned for optimal performance.
Architecture-based approaches \cite{xu2018reinforced,huang2019neural,razdaibiedina2023progressive} often take a different approach by learning separate sets of parameters dedicated to individual tasks. This enables the model to specialize and adapt its parameters for each task, avoiding interference between tasks and preserving task-specific knowledge. However, these approaches will introduce additional training parameters, which may not be very flexible and feasible for various LLMs. 

Therefore, we focus on rehearsal-based methods \cite{masson2019episodic,rolnick2019experience}, which are also called replay-based methods. These methods typically involve the storage of a subset of data from previous tasks. These stored data are used for future rehearsal through techniques such as experience replay \cite{rolnick2019experience} and representation consolidation \cite{bhat2022task}. Prior rehearsal-based approaches for language models mainly focus on using a little bit of precedent data \cite{scialom-etal-2022-fine,mok-etal-2023-large,zhang-etal-2023-citb}. However, these approaches often ignore discussion on real-world applications where previous data may be limited or unavailable. Although data-free knowledge distillation methods \cite{yin2020dreaming,smith2021dreaming} introduce auxiliary generative models for data construction, they are primarily designed for classification tasks, which may not be effective in LLMs, where a wide range of NLP tasks are involved. Additionally, similar to introducing teacher models \cite{miao-etal-2023-exploring,cheng-etal-2023-accelerating,huang2024response}, it can be challenging and time-consuming to train additional generative models. Self-distillation methods \cite{zhang2023exploring} may be useful, but catastrophic forgetting of the latest LLMs and the knowledge discrepancy among LLMs from distinct stages are still inevitable challenges.

In this work, we propose a rehearsal-based continual learning framework in which LLMs can be trained on self-synthesized data to retain the knowledge of the previous stages, with several demonstrations used during data construction. Unlike other approaches, our framework does not depend on additional generative models for data construction or require previous real data for rehearsal. This offers advantages in terms of data efficiency and application flexibility.

\begin{figure*}[tbp]
    \centering
    \includegraphics[width=\linewidth]{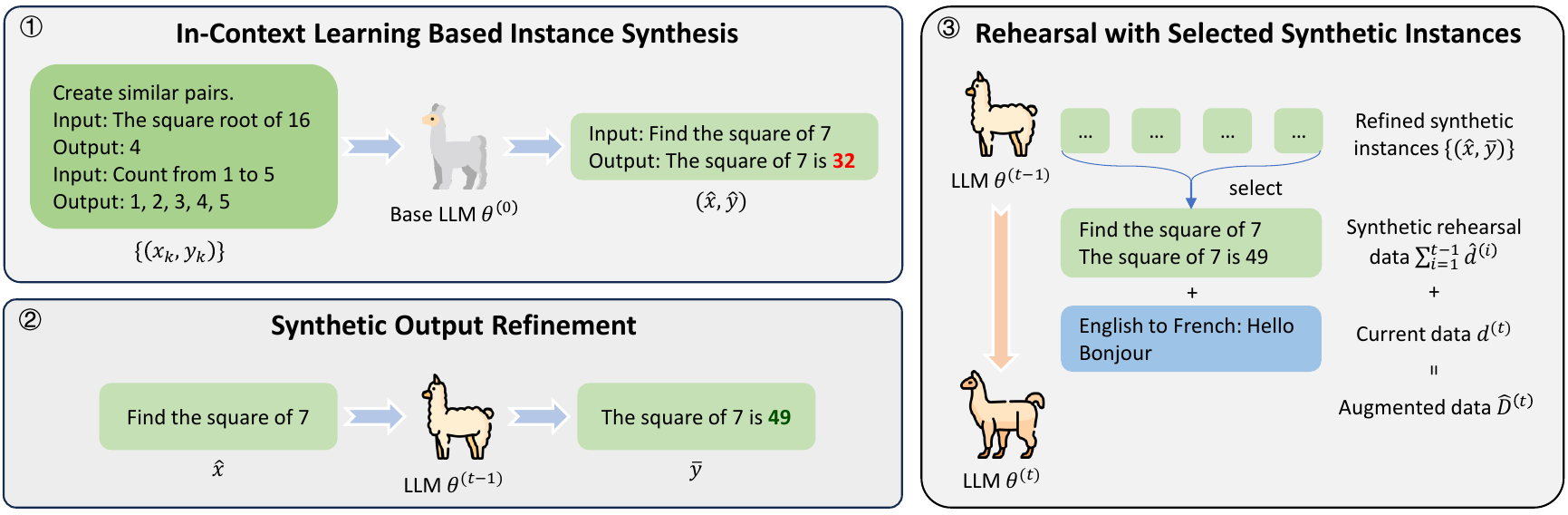}
    \caption{Our SSR framework. To mitigate catastrophic forgetting with limited or no rehearsal data, we first adopt the base LLM $\theta^{(0)}$ with in-context learning to generate synthetic instances $\{(\hat{x},\hat{y})\}$. We then utilize the latest LLM $\theta^{(t-1)}$ to generate the refined output $\bar{y}$ based on $\hat{x}$. Finally, diverse high-quality synthetic instances are selected for rehearsal in the future stages.}
    \label{fig:arch}
\end{figure*}

\section{Rehearsal-Based Continual Learning}

In continual learning, the LLM is sequentially updated for $T$ stages, with
each stage $t$ having its corresponding instruction data $d^{(t)}$. 
To mitigate catastrophic forgetting, in each stage $t$, 
rehearsal-based methods \cite{scialom-etal-2022-fine,mok-etal-2023-large} sample some training instances of previous stages to expand the training data in the current stage. Formally, the augmented training data $D^{(t)}$ can be formulated as follows:
\begin{equation}
    D^{(t)}=d^{(t)}\bigcup \sum_{i=1}^{t-1}(rd^{(i)}),
\end{equation}
where $r$ represents the rehearsal ratio determining the percentage of sampled training instances. 
Finally, we use $D^{(t)}$ to fine-tune the LLM $\theta^{(t-1)}$, obtaining the updated LLM $\theta^{(t)}$. 
Particularly, in the first stage, we fine-tune the base LLM $\theta^{(0)}$ on $D^{(1)}=d^{(1)}$. 
By doing so, the catastrophic forgetting problem of LLM can be effectively alleviated, which has been verified in previous studies \cite{scialom-etal-2022-fine,mok-etal-2023-large,zhang-etal-2023-citb}.

\section{Our Framework}
\label{sec:framework}

In this section, we detail the proposed \textbf{S}elf-\textbf{S}ynthesized \textbf{R}ehearsal (\textbf{SSR}) framework, which involves three main steps: 1) in-context learning based instance synthesis, 2) synthetic output refinement, and 3) rehearsal with selected synthetic instances, as illustrated in Figure~\ref{fig:arch}.

\paragraph{In-Context Learning Based Instance Synthesis}
Rehearsal-based methods utilize the training instances to cache the knowledge acquired by the LLM from previous stages.
Nevertheless, in real-world scenarios where a publicly-released LLM checkpoint is used, the availability of original training data may be limited.
To address this limitation, we try to generate rehearsal training instances synthetically.
To ask the LLM to follow abstract instructions, we leverage the in-context learning (ICL) capability of LLMs for instance synthesis. 

Formally, during each training stage $t$, we first acquire $K$ demonstrations $\{(x_k,y_k)\}_{k=1}^K$. To retain previously acquired knowledge, these demonstrations can be collected from the previous instruction data $d^{(t-1)}$ or manually constructed containing similar knowledge to $d^{(t-1)}$. We concatenate all demonstrations and utilize the base LLM to generate the synthetic instance $(\hat{x}, \hat{y})=\mathrm{LLM}(\mathrm{concat}_{k=1}^K(x_k,y_k);\theta^{(0)})$. 
By reordering the demonstrations and sampling multiple times, we can easily obtain different synthetic instances. It should be noted that we use the base LLM $\theta^{(0)}$ rather than the latest LLM $\theta^{(t-1)}$ to conduct ICL. This is because the ICL ability of LLMs tends to exhibit a significant degradation after supervised fine-tuning (SFT) on specific tasks, as analyzed in \cite{wang2023twostage}.

\paragraph{Synthetic Output Refinement} 
Through the above process, we obtain a series of synthetic instances, some of which, however, may be of low quality with unreliable outputs.
To address this issue, we use the latest LLM $\theta^{(t-1)}$ to refine the output of each synthetic instance:
$\bar{y}= \mathrm{LLM}(\hat{x};\theta^{(t-1)})$.
By doing so, we can ensure that each refined synthetic instance $(\hat{x}, \bar{y})$ retains the knowledge acquired by the latest LLM.

\paragraph{Rehearsal with Selected Synthetic Instances}
Finally, we select the refined synthetic instances for rehearsal. During this process, to ensure the diversity and quality of selected synthetic instances, we first adopt a clustering algorithm (e.g. K-means) to group $\{(\hat{x},\bar{y})\}$ into $C$ clusters. Then we calculate the distance between each synthetic instance and its corresponding cluster centroid, and finally select a certain amount of synthetic instances near cluster centroids as the rehearsal data.

Formally, we use $\hat{d}^{(t-1)}$ to represent the set of selected synthetic instances. Thus, the augmented training data in stage $t$ can be formulated as
\begin{equation}
    \hat{D}^{(t)}=d^{(t)}\bigcup \sum_{i=1}^{t-1}\hat{d}^{(i)},
\end{equation}
where $\hat{d}^{(i)}$ denotes the selected synthetic data similar to the previous training data $d^{(i)}$. 
Note that $\hat{d}^{(1)}, \hat{d}^{(2)}, ..., \hat{d}^{(t-2)}$ have been generated in previous stages, thus we will not regenerate them in stage $t$.
Finally, we use $\hat{D}^{(t)}$ to fine-tune $\theta^{(t-1)}$, updating the LLM as $\theta^{(t)}$.
In this way, the LLM can preserve previously learned knowledge even without real data from previous stages.

\section{Experiments}

\subsection{Setup}

\paragraph{Datasets}
We conduct several groups of experiments on the SuperNI dataset \cite{wang-etal-2022-super}, a vast and comprehensive benchmark dataset for instruction tuning. 
First, to simulate a typical continual learning process, we choose a subset of 10 tasks from SuperNI, encompassing various categories and domains. Each task is trained in a separate stage for empirical studies. For each task, we randomly sample 2,000 instances for training and 500 for evaluation.
Please refer to Appendix \ref{app:task-info} for the details of these tasks. To simplify the empirical study, we adopt default continual learning orders on \{5, 10\} SuperNI tasks: QA $\rightarrow$ QG $\rightarrow$ SA $\rightarrow$ Sum. $\rightarrow$ Trans. ($\rightarrow$ DSG $\rightarrow$ Expl. $\rightarrow$ Para. $\rightarrow$ PE $\rightarrow$ POS).

\paragraph{Base LLMs}
Our main experiments involve three base LLMs: Llama-2-7b \cite{touvron2023llama2}, Llama-2-7b-chat \cite{touvron2023llama2}, Alpaca-7b \cite{alpaca}.

\paragraph{Baselines}
We compare SSR with the following baselines:

\begin{itemize}
    \item \textbf{Multi-task Learning (MTL)}. This is the most commonly used baseline, where all tasks are trained simultaneously.    \item \textbf{Non-rehearsal}. It is a naive baseline that the LLM is fine-tuned with only the instruction data $d^{(t)}$ in each stage $t$.
    \item \textbf{RandSel($r$)} \cite{scialom-etal-2022-fine}. We randomly sample $r=$~\{1, 10\}\% of the original instruction data for each previous task. Note that as mentioned in \cite{scialom-etal-2022-fine}, the abilities of language models can be effectively preserved with $r=$~1\%.
    \item \textbf{KMeansSel($r$)}. Unlike the above approach, we first employ K-means clustering to group real instances into 20 clusters and then select $r=$~\{1, 10\}\% of instances with the highest similarities to the cluster centroids.
    Here we adopt SimCSE \cite{gao-etal-2021-simcse} to obtain instance representations before clustering.
\end{itemize}
\paragraph{Evaluation Metrics}
Due to the diversity and the open-ended sequence generation characteristic of SuperNI tasks, we adopt the ROUGE-L metric \cite{lin-2004-rouge} to evaluate the performance of LLM on each task.
This metric shows a good alignment with human evaluation, as demonstrated by \citet{wang-etal-2022-super}. Besides, we follow \citet{lopez2017gradient} to consider the following metrics based on ROUGE-L. Here $a^{(i)}_{j}$ denotes the ROUGE-L performance on the task $j$ in training stage $i$. 

\begin{itemize}
    \item \textbf{Average ROUGE-L (AR)}. It is used to quantify the final average performance of LLM across all $T$ tasks in stage $T$, which is defined as follows:
    \begin{equation}
        \mathbf{AR}=\frac{1}{T}\sum_{i=1}^{T}a^{(T)}_{i}.
    \end{equation}
    \item \textbf{Forward Transfer (FWT)}. It evaluates the LLM's generalization ability on unseen tasks, measuring the average zero-shot performance $a^{(i-1)}_{i}$ on the next task $i$ in each stage $i-1$:
    \begin{equation}
        \mathbf{FWT}=\frac{1}{T-1}\sum_{i=2}^{T}a^{(i-1)}_{i}.
    \end{equation}
    \item \textbf{Backward Transfer (BWT)}. It is a metric used to evaluate the impact of learning subsequent tasks on a previous task. 
    For each task $i$ except for the final one, it compares the final performance $a^{(T)}_{i}$ to the online performance $a^{(i)}_{i}$ in stage $i$:
    \begin{equation}
        \mathbf{BWT}=\frac{1}{T-1}\sum_{i=1}^{T-1}(a^{(T)}_{i}-a^{(i)}_{i}).
    \end{equation}
    A negative BWT indicates that the LLM has forgotten some previously acquired knowledge. 
\end{itemize}

\begin{table*}[ht]
\centering
\small
\tabcolsep=11pt
\begin{tabular}{lcccccccccccc}
\toprule
\multirow{2}{*}{\textbf{Model}}
& \multicolumn{2}{c}{\textbf{Order 1}} & \multicolumn{2}{c}{\textbf{Order 2}} & \multicolumn{2}{c}{\textbf{Order 3}} & \multicolumn{2}{c}{\textbf{Avg.}} \\
\cmidrule(lr){2-3} \cmidrule(lr){4-5} \cmidrule(lr){6-7} \cmidrule(lr){8-9}
&  \textbf{AR} & \textbf{BWT} &  \textbf{AR} & \textbf{BWT} &  \textbf{AR} & \textbf{BWT} & \textbf{AR} & \textbf{BWT} \\
\midrule
\multicolumn{9}{c}{\textit{Llama-2-7b}} \\
\midrule
MTL & 53.05 & - & 53.05 & - & 53.05 & - & 53.05  & -\\
Non-rehearsal & 17.67 & -44.09 & 15.25 & -47.09 & 24.16 & -35.99 & 19.03  & -42.39 \\
\hdashline
RandSel(1\%) & 51.16 & -2.34 & 49.21 & -4.36 & 48.63 & -5.37 & 49.67  & -4.02 \\
KMeansSel(1\%) & 50.20 & -3.12 & 49.75 & -4.11 & 50.12 & -3.61 & 50.02  & -3.61 \\
RandSel(10\%) & 50.81 & -2.32 & 50.04 & -3.31 & 50.11 & -3.42 & 50.32  & -3.02 \\
KMeansSel(10\%) & 50.44 & -3.03 & 50.61 & -2.32 & 49.89 & -3.53 & 50.31  & -2.96 \\
\hdashline
SSR & \textbf{52.61} & \textbf{-0.23} & \textbf{51.70} & \textbf{-1.22} & \textbf{52.16} & \textbf{-0.93} & \textbf{52.16}  & \textbf{-0.79} \\
\midrule
\multicolumn{9}{c}{\textit{Llama-2-7b-chat}} \\
\midrule
MTL & 52.81 & - & 52.81 & - & 52.81 & - & 52.81  & -\\
Non-rehearsal & 23.87 & -36.31 & 30.96 & -27.41 & 42.06 & -13.50 & 32.30  & -25.74 \\
\hdashline
RandSel(1\%) & 51.28 & -1.96 & 49.77 & -3.70 & 49.41 & -4.29 & 50.15  & -3.32 \\
KMeansSel(1\%) & 51.82 & -1.25 & 50.71 & -2.44 & 50.22 & -3.42 & 50.92  & -2.37 \\
RandSel(10\%) & 50.59 & -2.57 & 50.72 & -2.45 & 50.24 & -2.87 & 50.52  & -2.63 \\
KMeansSel(10\%) & 50.81 & -2.55 & 51.39 & -1.42 & 50.22 & -2.84 & 50.81  & -2.27 \\
\hdashline
SSR & \textbf{52.52} & \textbf{-0.23} & \textbf{52.49} & \textbf{-0.35} & \textbf{52.73} & \textbf{0.05} & \textbf{52.58}  & \textbf{-0.18} \\
\midrule
\multicolumn{9}{c}{\textit{Alpaca-7b}} \\
\midrule
MTL & 52.79 & - & 52.79 & - & 52.79 & - & 52.79  & -\\
Non-rehearsal & 17.24 & -44.21 & 45.40 & -9.03 & 35.60 & -21.45 & 32.75  & -24.90 \\
\hdashline
RandSel(1\%) & 51.61 & -0.93 & 49.08 & -4.68 & 49.01 & -4.85 & 49.90  & -3.49 \\
KMeansSel(1\%) & 51.37 & -1.53 & 50.53 & -2.68 & 50.15 & -3.17 & 50.68  & -2.46 \\
RandSel(10\%) & 50.91 & -1.82 & 50.88 & -2.11 & 49.98 & -3.59 & 50.59  & -2.51 \\
KMeansSel(10\%) & 50.78 & -2.05 & 51.20 & -1.76 & 49.76 & -3.48 & 50.58  & -2.43 \\
\hdashline
SSR & 
\textbf{52.52} & \textbf{-0.14} & \textbf{51.74} & \textbf{-1.21} & \textbf{52.33} & \textbf{-0.51} & \textbf{52.20} & \textbf{-0.62} \\
\bottomrule
\end{tabular}
\caption{Final results on 5 SuperNI tasks under different continual learning (CL) orders. For more details, please refer to Appendix \ref{app:five-task}.}
\label{tab:five-task}
\end{table*}

\paragraph{Implementation Details}
During training, we utilize LoRA \cite{hu2021lora} with query and value projection matrices in the self-attention module to train LLMs, setting the LoRA rank to 8 and the dropout rate of 0.1. We employ the Adam optimizer with an initial learning rate of 2e-4. The global batch size is 32 for all our experiments. Besides, we set the maximum length of the input to 1,024 and the counterpart of the output to 512. Following \citet{luo2023empirical}, we train each LLM for 3 epochs and use the final checkpoint for evaluation.

To conduct ICL, we utilize 1\% of the training data from SuperNI tasks as demonstrations, considering $K=$~2 demonstrations and sampling multiple times to obtain diverse synthetic instances. When clustering instances, we use K-means clustering with $C=$~20 clusters for synthetic instances of SuperNI tasks, which is similar to KMeansSel($r$).

\subsection{Experiments on 5 SuperNI Tasks}
\label{subsec:five-task}

Table \ref{tab:five-task} presents the experimental results on 5 SuperNI tasks. Overall, regardless of the continual learning order and the base LLM, 
SSR consistently outperforms all rehearsal-based baselines, exhibiting an improvement of approximately 2 scores in both the AR and BWT metrics. This result shows the superiority of SSR in mitigating catastrophic forgetting. 
Particularly, SSR closely approaches MTL which sets the upper bound of the AR performance.
Besides,
compared to rehearsal-based baselines, RandSel($r$) and KMeansSel($r$), SSR is more data-efficient 
with only 1\% real data utilization for ICL and only synthetic data of previous stages for rehearsal. 

After further analysis, we draw the following conclusions:

\paragraph{Non-rehearsal vs. rehearsal}
The non-rehearsal baseline shows the poorest, indicating severe catastrophic forgetting. Besides, it exhibits the highest metric variance, signifying its lack of robustness in different CL orders.
In contrast, SSR and rehearsal-based baselines maintain better and more consistent performance regardless of the CL order. 

\begin{figure*}[ht]
    \centering    \includegraphics[width=\linewidth]{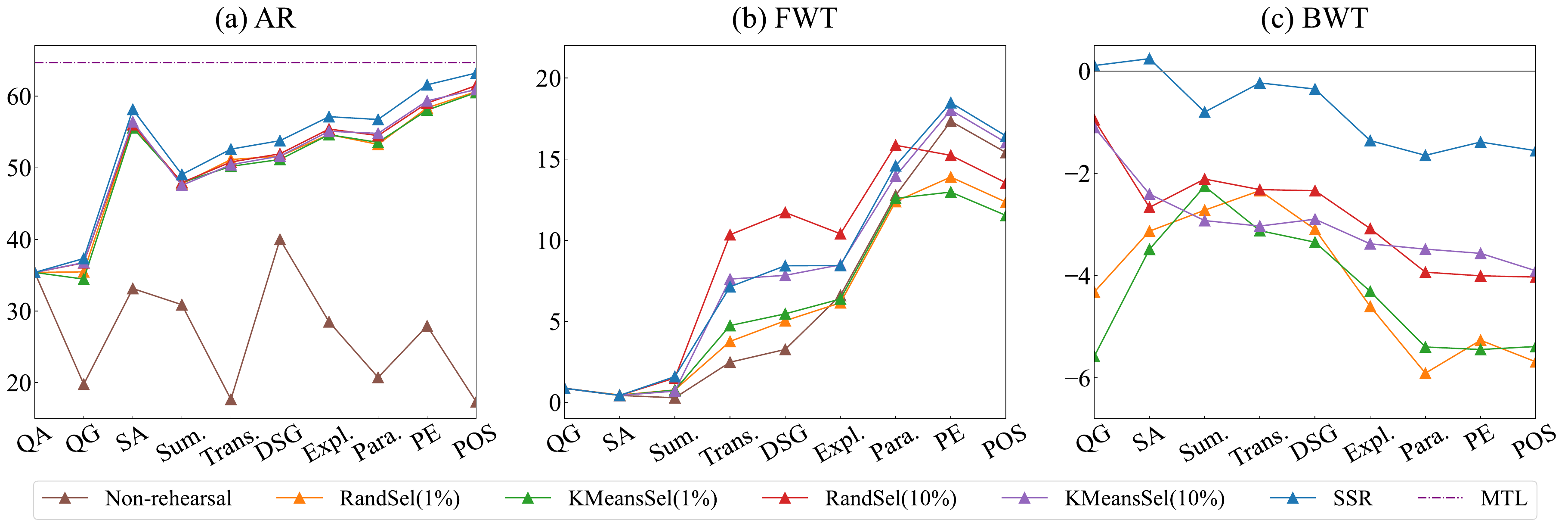}
    \caption{AR, FWT, and BWT during continual learning for Llama-2-7b on 10 SuperNI tasks.}
    \label{fig:overall}
\end{figure*}

\paragraph{Effect of $r$}
The appropriate rehearsal ratio $r$ varies depending on the continual learning orders. A higher $r$ is beneficial in certain cases, as observed in CL orders 2 and 3. However, this is not always the case. In CL order 1, regardless of the instance sampling strategy employed, the rehearsal-based baselines with $r=$~1\% consistently outperform their $r=$~10\% counterparts, respectively.

\paragraph{RandSel($r$) vs. KMeansSel($r$)}
When comparing RandSel($r$) and KMeansSel($r$), we can observe that K-means clustering-based selection of previous data for rehearsal may slightly enhance the model performance when using only $r=$ 1\%, demonstrating the importance of data representativeness. However, when $r$ is set to 10\%, significant differences in model performance may not be observed for LLMs such as Llama-2-7b and Alpaca-7b.

\begin{table}[t]
\centering
\small
\tabcolsep=8pt
\begin{tabular}{lccc}
\toprule
\textbf{Model}& \textbf{AR} & \textbf{FWT} & \textbf{BWT} \\
\midrule
\multicolumn{4}{c}{\textit{Llama-2-7b}} \\\midrule
MTL
& 64.69 & - & - \\
Non-rehearsal 
& 17.33 & 15.41 & -53.64 \\
\hdashline
RandSel(1\%)
& 60.64 & 12.35 & -5.69 \\
KMeansSel(1\%) & 60.51 & 11.53 & -5.39 \\
RandSel(10\%)
& 61.49 & 13.54 & -4.03 \\
KMeansSel(10\%) & 60.93 & 16.03 & -3.90 \\
\hdashline
SSR
& \textbf{63.23} & \textbf{16.43} & \textbf{-1.56} \\
\bottomrule
\end{tabular}
\caption{Final results for Llama-2-7b on 10 SuperNI tasks.}
\label{tab:ten-task}
\end{table}

\subsection{Experiments on 10 SuperNI Tasks}
\label{subsec:ten-task}

To further investigate the effectiveness of SSR in longer continual learning sequences, we evaluate SSR and all baselines on 10 SuperNI tasks. Table \ref{tab:ten-task} shows that SSR surpasses all rehearsal-based baselines across all metrics. 
Moreover, as illustrated in Figure~\ref{fig:overall}, SSR consistently achieves better performance in terms of AR and BWT compared to rehearsal-based baselines throughout the entire continual learning process. Although SSR with Llama-2-7b as the base LLM falls behind RandSel(10\%) in terms of FWT in the early stage, it gradually strengthens its performance as the number of training stages increases, eventually surpassing RandSel(10\%). Please refer to Appendix \ref{app:ten-task} for more details.

\subsection{Experiments on the Generalization Capability Preservation of Alpaca-7b}
\label{subsec:alpaca-52k}

To further analyze the preservation of LLM's generalization ability in a broader domain beyond SuperNI tasks, we utilize Alpaca-7b to conduct continual learning on 5 SuperNI tasks and then investigate whether SSR can preserve the abilities of Alpaca-7b gained from the Alpaca-52k dataset\footnote{\url{https://huggingface.co/datasets/tatsu-lab/alpaca}}. Here, Llama-7b \cite{touvron2023llama} serves as the base LLM $\theta^{(0)}$, and Alpaca-7b is considered as the updated LLM $\theta^{(1)}$ after fine-tuning on Alpaca-52k. Therefore, we also generate synthetic instances similar to Alpaca-52k for SSR and use Alpaca-52k for rehearsal-based baselines.

We evaluate the LLM from three perspectives: 1) General instruction-following ability.
We use AlpacaEval~2.0\footnote{\url{https://github.com/tatsu-lab/alpaca_eval}} as an automatic evaluator. 
Concretely, we measure the LLM's performance in terms of the \textbf{win rate}, comparing the LLM's generations with those generated by \texttt{gpt-4-turbo}. To minimize financial costs, we utilize ChatGPT as the evaluation annotator.
2) General language understanding ability. We leverage the MMLU \cite{hendrycks2021measuring} benchmark, where \textbf{accuracy (Acc.)} is used as the evaluation metric. 3) Task-specific ability. We evaluate the AR performance of the LLM on 5 SuperNI tasks.
Please refer to Appendix \ref{app:alpaca-exp} for more details.

\begin{table}[t]
\centering
\small
\begin{tabular}{lcccc}
\toprule
\multirow{2}{*}{\textbf{Model}}    & \textbf{AlpacaEval}  & {\textbf{MMLU}} & \textbf{SuperNI} \\
\cmidrule(lr){2-2}  \cmidrule(lr){3-3} \cmidrule(lr){4-4} 
& win rate &  Acc. & AR \\
\midrule
Alpaca-7b & 21.08 & 41.0 & 22.80 \\
Non-rehearsal & 8.82 & 34.1 & 17.24 \\
\hdashline
RandSel(1\%) & 19.45 & 36.9 & 51.80 \\
RandSel(10\%) & \textbf{20.06} & 36.4 & 50.47 \\
\hdashline
SSR & 19.68 & \textbf{37.1}  & \textbf{52.11} \\
\bottomrule
\end{tabular}
\caption{Final results on Alpaca-52k + 5 SuperNI tasks.}
\label{tab:alpaca-52k}
\end{table}

From Table \ref{tab:alpaca-52k}, we observe that SSR not only achieves the best on the 5 newly learned tasks but also maintains comparable or even superior performance on AlpacaEval and MMLU. 
These findings suggest that SSR effectively preserves the generalization ability of Alpaca-7b throughout the continual learning process, even in the absence of Alpaca-52k as rehearsal data. This highlights the great potential of SSR in general domains.

\subsection{Analysis}

\paragraph{Effect of in-context learning} To investigate the effect of in-context learning on SSR, we conduct experiments for Llama-2-7b on 5 SuperNI tasks, introducing the following variants: (a) \textbf{Llama-2-7b$\Rightarrow$Llama-7b}. This variant validates the scenario where we acquire a public-released fine-tuned LLM checkpoint, but the original base LLM is unavailable. Thus we employ a different LLM Llama-7b to conduct ICL.  (b) \textbf{Llama-2-7b$\Rightarrow$Alpaca-7b}. Similar to the above one, but using Alpaca-7b. (c) \textbf{train demos$\Rightarrow$new demos}. In this variant, we utilize demonstrations that are not included in the previous training data but belong to the same SuperNI task, simulating manually constructed demonstrations to conduct ICL. (d) \textbf{w/ input-only demos}. 
This variant utilizes only the instance inputs in previous stages as demonstrations, simulating the scenario where real instances lack output annotations.

\begin{table}[]
\centering
\tabcolsep=8pt
\small
\begin{tabular}{lcccccccccccc}
\toprule
\textbf{Model} & \textbf{AR} & \textbf{BWT} \\
\midrule
\multicolumn{3}{c}{\textit{Llama-2-7b}} \\
\midrule
SSR & 52.61 & \textbf{-0.23} \\
~~Llama-2-7b$\Rightarrow$Llama-7b & \textbf{52.71} & -0.36 \\
~~Llama-2-7b$\Rightarrow$Alpaca-7b & 52.07 & -0.78 \\
~~train demos$\Rightarrow$new demos & 52.54 & -0.44 \\
~~w/ input-only demos & 52.61 & -0.34 \\
\bottomrule
\end{tabular}
\caption{Effect of in-context learning for Llama-2-7b on 5 SuperNI tasks.}
\label{tab:five-task-icl}
\end{table}

Table~\ref{tab:five-task-icl} illustrates that SSR can perform well even without the original base LLM or demonstrations from previous training data to conduct ICL. This provides convenience for replacing some ICL components in practical application scenarios.
Comparing SSR and its variants (a) to (b), we notice a slight decrease in performance when conducting ICL using Alpaca-7b. This highlights the limitation of this fine-tuned LLM in terms of ICL capability.
Besides, ICL with input-only demonstrations also yields comparable performance, indicating that output annotations are also not essential for ICL, further verifying the robustness of SSR.

\begin{figure}
    \centering \includegraphics[width=\linewidth]{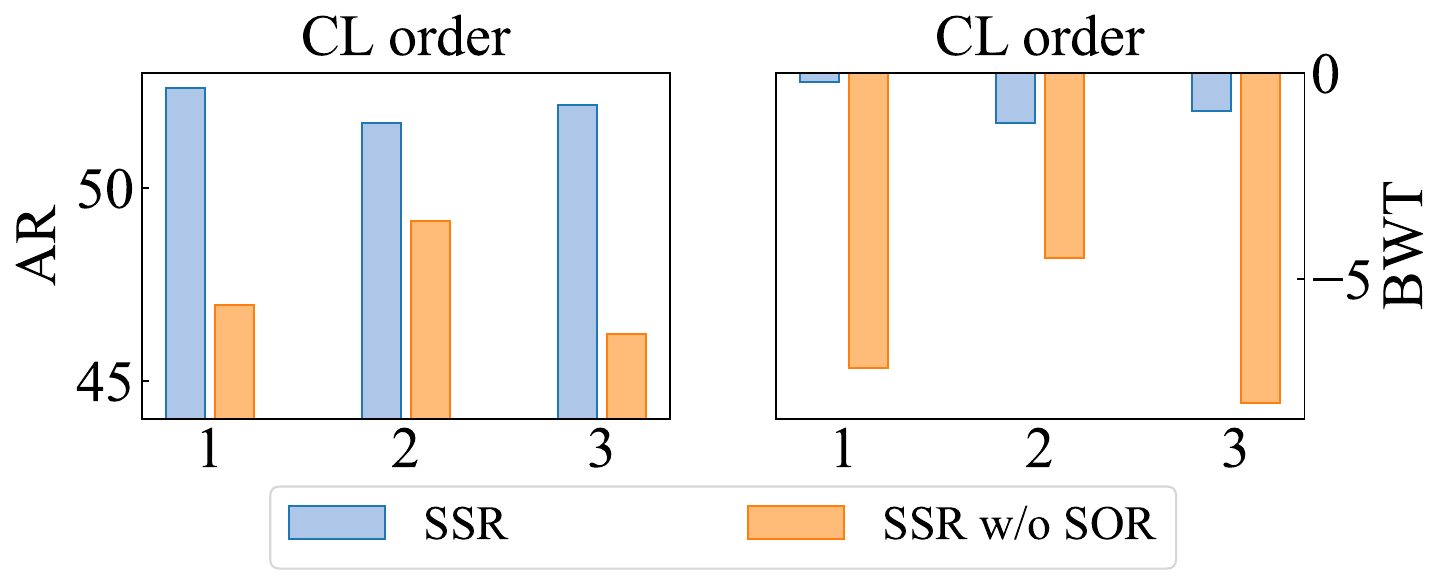}
    \caption{Effect of synthetic output refinement (SOR) for Llama-2-7b on 5 SuperNI tasks under different continual learning orders.}
    \label{fig:kd}
\end{figure}

\paragraph{Effect of synthetic output refinement} 
In Section \ref{sec:framework}, we claim that synthetic output refinement provides more reliable synthetic outputs from the latest LLM. 
To verify the effectiveness, we conduct an experiment where SSR is implemented without synthetic output refinement. 

As illustrated in Figure~\ref{fig:kd}, this results in lower AR values and significant BWT inferiority, highlighting the negative impact of data noise originating from the base LLM. In contrast, by incorporating the synthetic input with the refined output, SSR can maintain the predictive distribution of the latest LLM during rehearsal, preserving the acquired knowledge.

\begin{figure}[t]
    \centering \includegraphics[width=\linewidth]{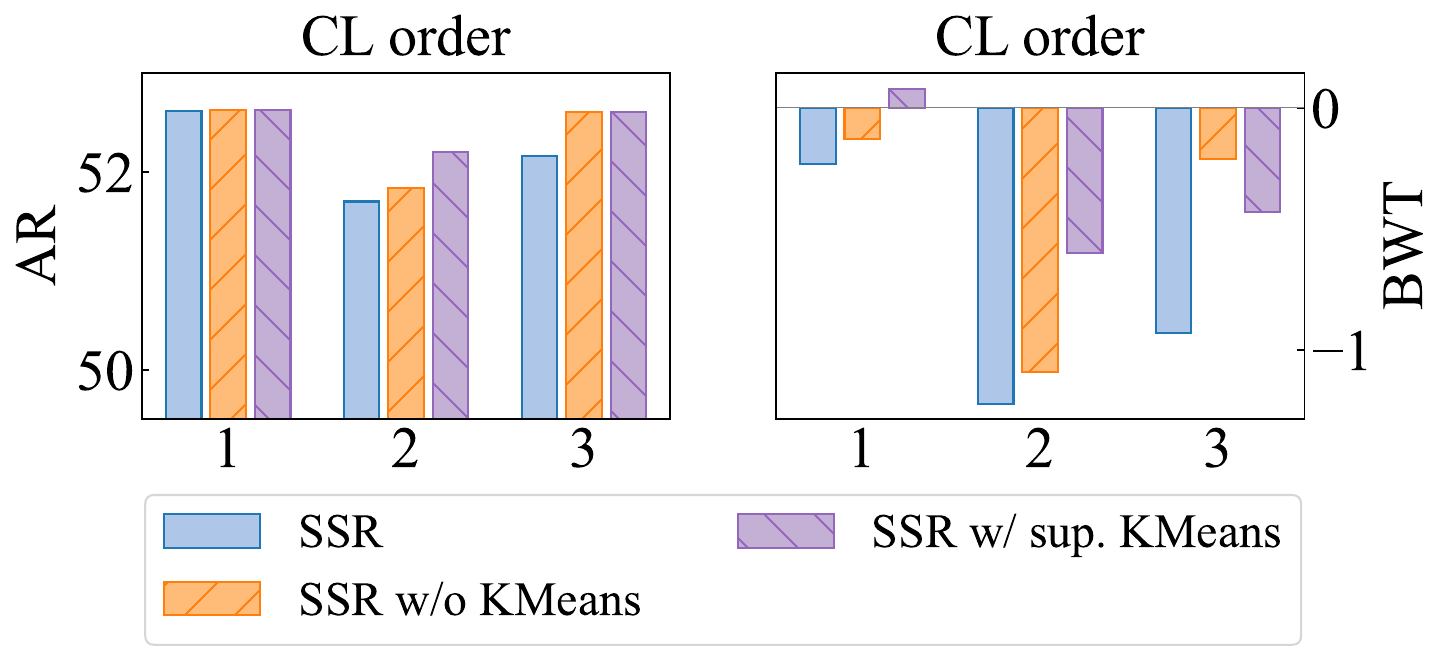}
    \caption{Effect of K-means clustering for Llama-2-7b on 5 SuperNI tasks under different continual learning orders. }
    \label{fig:sup-kmeans}
\end{figure}

\paragraph{Effect of K-means clustering on synthetic instance selection}
In terms of application flexibility, we utilize an unsupervised K-means clustering algorithm, fitting and predicting solely with synthetic instances.  
To explore the effect of K-means clustering, we compare SSR with the following variants: (a) \textbf{SSR~w/o~KMeans}: Random selection of synthetic instances. (b) \textbf{SSR~w/~sup.~KMeans}: K-means clustering-based synthetic instance selection, with the supervision of real instances to fit the K-means clustering and then predict on synthetic instances.

Figure~\ref{fig:sup-kmeans} shows that the supervised K-means clustering method leads to a slight improvement in AR and reduces forgetting with larger BWT. Thus, incorporating real instances during clustering may allow for a more representative selection. Nonetheless, clustering is not essential in the absence of supervision, because SSR with random selection for synthetic instances can outperform SSR, sometimes even surpass SSR with supervised K-means. This indicates a certain level of robustness of SSR.

\begin{figure}[t]
    \centering \includegraphics[width=\linewidth]{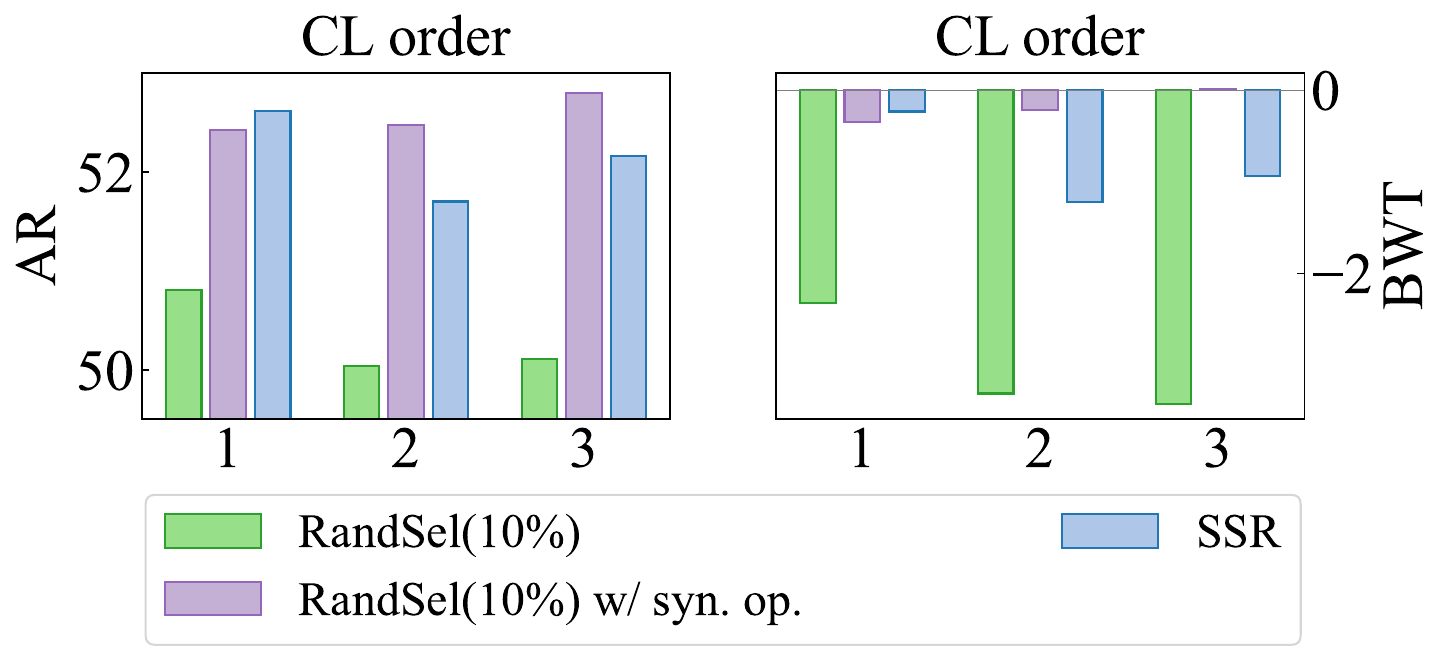}
    \caption{Effect of synthetic inputs and outputs for Llama-2-7b on 5 SuperNI tasks under different continual learning orders. Note that RandSel(10\%) w/ syn. op. (synthetic outputs) in CL order 3 has the best BWT value of 0.02.}
    \label{fig:real-input}
\end{figure}

\begin{figure}[t]
    \centering \includegraphics[width=\linewidth]{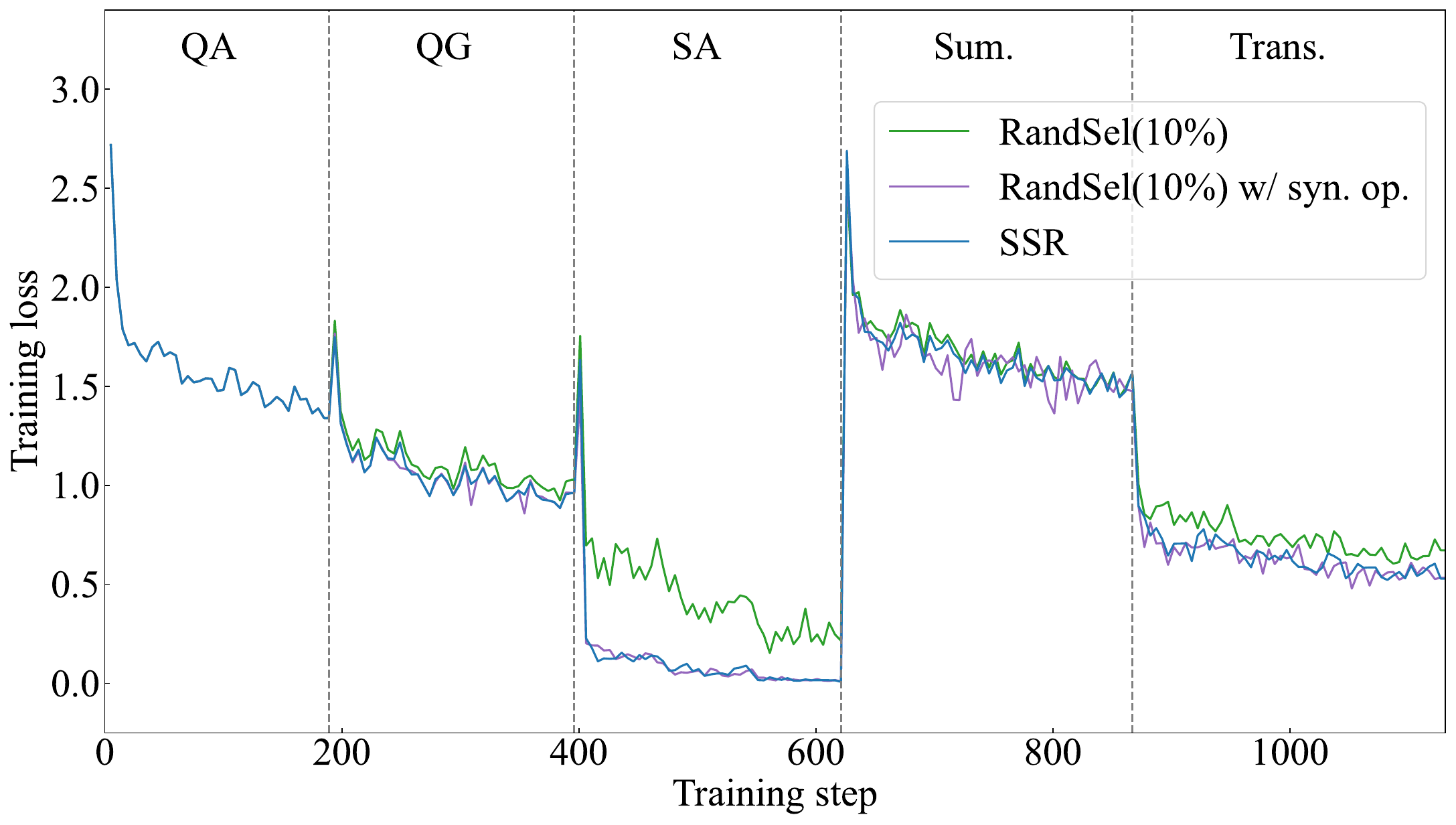}
    \caption{Effect of synthetic inputs and outputs on loss curve for Llama-2-7b on 5 SuperNI tasks.}
    \label{fig:loss}
\end{figure}

\paragraph{Real instances vs. synthetic instances} Our main experiments demonstrate surprising results that rehearsal with synthetic instances may surpass those with real instances.
For the comparison of real and synthetic instances, we consider the following variants: (a) \textbf{RandSel (10\%)}: Real inputs and outputs for rehearsal. (b) \textbf{RandSel (10\%)~w/~syn. op.}: Real inputs and synthetic outputs for rehearsal. Concretely, we regenerate the outputs of randomly sampled previous instances by the latest LLM, with similar operations to SSR.
Figure~\ref{fig:real-input} demonstrates that RandSel (10\%) with real inputs and synthetic outputs for rehearsal, outperforms RandSel (10\%) utilizing only real instances. Meanwhile, SSR, which leverages both synthetic inputs and outputs for rehearsal, achieves intermediate performance, sometimes even surpassing the other two. This indicates that real instances are not always essential and appropriate for the continual learning of LLMs. As depicted in Figure~\ref{fig:loss}, real instances often lead to a slower descent in loss. Therefore, they may not be conducive to optimization due to the distribution gap between distinct datasets. Conversely, synthetic instances, with lower model perplexity, embody the LLM's real-time acquired knowledge, which aids in smoothing the data distribution and discovering better local optima for the LLM.

\begin{figure}[t]
    \centering \includegraphics[width=\linewidth]{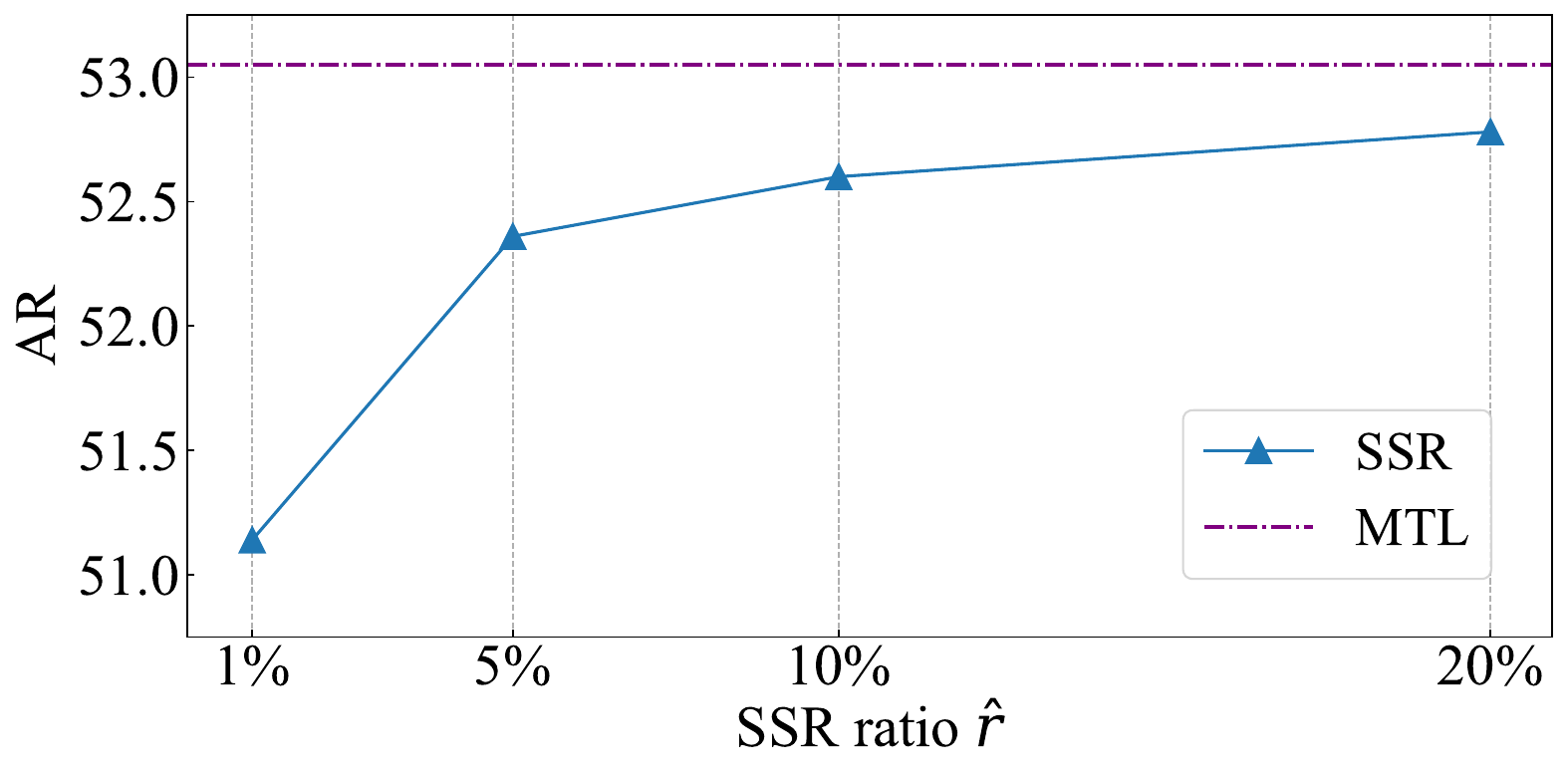}
    \caption{Effect of the SSR ratio $\hat{r}$ for Llama-2-7b on 5 SuperNI tasks.}
    \label{fig:data-ratio}
\end{figure}

\paragraph{Effect of synthetic instance quantity}
Here, we define $\hat{r}=|\hat{d}^{(i)}|/|d^{(i)}|$ as the \textbf{SSR ratio}, which represents the proportion of selected synthetic data compared to the original training data size. By default, we retain synthetic instances at an SSR ratio of $\hat{r}=$ 10\%.
However, as depicted in Figure~\ref{fig:data-ratio}, increasing $\hat{r}$ can lead to further improvements in the final AR, highlighting the potential of SSR. Moreover, it is important to note that the training cost and memory limit should also be taken into consideration when determining an appropriate $\hat{r}$.

\paragraph{SSR vs. regularization-based and architecture-based methods} In this paper, we focus on generating synthetic rehearsal data for instruction tuning. Prior work \cite{zhang-etal-2023-citb} has demonstrated that rehearsal-based approaches are generally superior to regularization-based and architecture-based ones for instruction tuning of language models. Table \ref{tab:ssr-vs-reg} presents experimental results using two classical regularization-based baselines (L2 and EWC) for Llama-2-7b under CL order 1, with SSR still demonstrating its superiority. Besides, these lightweight strategies can be easily combined with SSR, potentially leading to further improvements in model performance. Moreover, architecture-based approaches heavily rely on additional task-specific parameters, which may not be practical in real-world applications where the inference time of LLMs is a crucial consideration.

\begin{table}[t]
\centering
\small
\tabcolsep=8pt
\begin{tabular}{lccc}
\toprule
\textbf{Model}
& \textbf{AR} 
& \textbf{FWT}
& \textbf{BWT} \\
\midrule
\multicolumn{4}{c}{\textit{Llama-2-7b}} \\
\midrule
MTL
& 53.05 & - & - \\
Non-rehearsal 
& 17.67 & 2.49 & -44.09 \\
\hdashline
\textit{regularization-based} \\
L2 & 29.22 
& 7.08 
& -28.45 \\
EWC & 15.94 
& 2.43
& -46.30 \\
\hdashline
SSR & \textbf{52.61} 
& \textbf{7.14}
& \textbf{-0.23} \\
\bottomrule
\end{tabular}
\caption{Comparison between SSR and regularization-based methods for Llama-2-7b on 5 SuperNI tasks under CL order 1.}
\label{tab:ssr-vs-reg}
\end{table}

\section{Conclusion}

In this work, we propose Self-Synthesized Rehearsal (SSR), a continual learning framework for mitigating catastrophic forgetting in LLMs, to effectively preserve knowledge without relying on real data during rehearsal. Through extensive experiments, SSR demonstrates its data efficiency and superior performance to conventional rehearsal-based approaches. Besides, it preserves LLM's generalization capability both in specific and general domains, with flexibility and robustness in real-world scenarios. 
Overall, SSR presents a promising solution for continual learning of LLMs in real-world settings, with implications for maintaining the acquired abilities of LLMs.

\section*{Limitations}

Although SSR demonstrates superior performance in terms of AR and BWT, it may not always achieve the best FWT score, as shown in Figure \ref{fig:overall}(b). However, as discussed in Subsection \ref{subsec:alpaca-52k}, SSR effectively preserves the generalization capabilities of LLMs in general domains, highlighting its practical value. For the final FWT results on the 5 SuperNI tasks, please refer to Table \ref{tab:five-task-fwt} in Appendix \ref{app:five-task}.  Additionally, synthetic instances generated by LLMs may potentially contain unsafe content due to data bias during training.

\section*{Acknowledgements}

The project was supported by National Key R\&D Program of China (No. 2022ZD0160501), National Natural Science Foundation of China (No. 62276219), and the Public Technology Service Platform Project of Xiamen (No.3502Z20231043). We also thank the reviewers for their insightful comments.

\bibliography{anthology,custom}

\begin{thebibliography}{32}
\expandafter\ifx\csname natexlab\endcsname\relax\def\natexlab#1{#1}\fi

\bibitem[{Bhat et~al.(2022)Bhat, Zonooz, and Arani}]{bhat2022task}
Prashant Bhat, Bahram Zonooz, and Elahe Arani. 2022.
\newblock \href {http://arxiv.org/abs/2207.06267} {Task agnostic representation consolidation: a self-supervised based continual learning approach}.

\bibitem[{Cha et~al.(2021)Cha, Hsu, Hwang, Calmon, and Moon}]{cha2021cpr}
Sungmin Cha, Hsiang Hsu, Taebaek Hwang, Flavio~P. Calmon, and Taesup Moon. 2021.
\newblock \href {http://arxiv.org/abs/2006.07326} {Cpr: Classifier-projection regularization for continual learning}.

\bibitem[{Cheng et~al.(2023)Cheng, Zhu, Xu, Li, Li, and Zou}]{cheng-etal-2023-accelerating}
Xuxin Cheng, Zhihong Zhu, Wanshi Xu, Yaowei Li, Hongxiang Li, and Yuexian Zou. 2023.
\newblock \href {https://doi.org/10.18653/v1/2023.findings-emnlp.597} {Accelerating multiple intent detection and slot filling via targeted knowledge distillation}.
\newblock In \emph{Findings of the Association for Computational Linguistics: EMNLP 2023}, pages 8900--8910, Singapore. Association for Computational Linguistics.

\bibitem[{de~Masson~d\textquotesingle Autume et~al.(2019)de~Masson~d\textquotesingle Autume, Ruder, Kong, and Yogatama}]{masson2019episodic}
Cyprien de~Masson~d\textquotesingle Autume, Sebastian Ruder, Lingpeng Kong, and Dani Yogatama. 2019.
\newblock \href {https://proceedings.neurips.cc/paper_files/paper/2019/file/f8d2e80c1458ea2501f98a2cafadb397-Paper.pdf} {Episodic memory in lifelong language learning}.
\newblock In \emph{Advances in Neural Information Processing Systems}, volume~32. Curran Associates, Inc.

\bibitem[{Gao et~al.(2021)Gao, Yao, and Chen}]{gao-etal-2021-simcse}
Tianyu Gao, Xingcheng Yao, and Danqi Chen. 2021.
\newblock \href {https://doi.org/10.18653/v1/2021.emnlp-main.552} {{S}im{CSE}: Simple contrastive learning of sentence embeddings}.
\newblock In \emph{Proceedings of the 2021 Conference on Empirical Methods in Natural Language Processing}, pages 6894--6910, Online and Punta Cana, Dominican Republic. Association for Computational Linguistics.

\bibitem[{Hendrycks et~al.(2021)Hendrycks, Burns, Basart, Zou, Mazeika, Song, and Steinhardt}]{hendrycks2021measuring}
Dan Hendrycks, Collin Burns, Steven Basart, Andy Zou, Mantas Mazeika, Dawn Song, and Jacob Steinhardt. 2021.
\newblock \href {http://arxiv.org/abs/2009.03300} {Measuring massive multitask language understanding}.

\bibitem[{Hu et~al.(2021)Hu, Shen, Wallis, Allen-Zhu, Li, Wang, Wang, and Chen}]{hu2021lora}
Edward~J. Hu, Yelong Shen, Phillip Wallis, Zeyuan Allen-Zhu, Yuanzhi Li, Shean Wang, Lu~Wang, and Weizhu Chen. 2021.
\newblock \href {http://arxiv.org/abs/2106.09685} {Lora: Low-rank adaptation of large language models}.

\bibitem[{Huang et~al.(2024)Huang, Wang, Gao, Song, and Su}]{huang2024response}
Jianheng Huang, Ante Wang, Linfeng Gao, Linfeng Song, and Jinsong Su. 2024.
\newblock \href {https://doi.org/10.1609/aaai.v38i16.29790} {Response enhanced semi-supervised dialogue query generation}.
\newblock In \emph{Proceedings of the AAAI Conference on Artificial Intelligence}, volume~38, pages 18307--18315.

\bibitem[{Huang et~al.(2019)Huang, François-Lavet, and Rabusseau}]{huang2019neural}
Shenyang Huang, Vincent François-Lavet, and Guillaume Rabusseau. 2019.
\newblock \href {http://arxiv.org/abs/1909.06686} {Neural architecture search for class-incremental learning}.

\bibitem[{Huang et~al.(2021)Huang, Zhang, Chen, Wang, and Yang}]{huang2021continual}
Yufan Huang, Yanzhe Zhang, Jiaao Chen, Xuezhi Wang, and Diyi Yang. 2021.
\newblock \href {http://arxiv.org/abs/2104.05489} {Continual learning for text classification with information disentanglement based regularization}.

\bibitem[{Kirkpatrick et~al.(2017)Kirkpatrick, Pascanu, Rabinowitz, Veness, Desjardins, Rusu, Milan, Quan, Ramalho, Grabska-Barwinska, Hassabis, Clopath, Kumaran, and Hadsell}]{Kirkpatrick_2017}
James Kirkpatrick, Razvan Pascanu, Neil Rabinowitz, Joel Veness, Guillaume Desjardins, Andrei~A. Rusu, Kieran Milan, John Quan, Tiago Ramalho, Agnieszka Grabska-Barwinska, Demis Hassabis, Claudia Clopath, Dharshan Kumaran, and Raia Hadsell. 2017.
\newblock \href {https://doi.org/10.1073/pnas.1611835114} {Overcoming catastrophic forgetting in neural networks}.
\newblock \emph{Proceedings of the National Academy of Sciences}, 114(13):3521–3526.

\bibitem[{Li et~al.(2022)Li, Chen, Cho, Hao, Liu, Xing, Guo, and Liu}]{li-etal-2022-overcoming}
Dingcheng Li, Zheng Chen, Eunah Cho, Jie Hao, Xiaohu Liu, Fan Xing, Chenlei Guo, and Yang Liu. 2022.
\newblock \href {https://doi.org/10.18653/v1/2022.naacl-main.398} {Overcoming catastrophic forgetting during domain adaptation of seq2seq language generation}.
\newblock In \emph{Proceedings of the 2022 Conference of the North American Chapter of the Association for Computational Linguistics: Human Language Technologies}, pages 5441--5454, Seattle, United States. Association for Computational Linguistics.

\bibitem[{Lin(2004)}]{lin-2004-rouge}
Chin-Yew Lin. 2004.
\newblock \href {https://aclanthology.org/W04-1013} {{ROUGE}: A package for automatic evaluation of summaries}.
\newblock In \emph{Text Summarization Branches Out}, pages 74--81, Barcelona, Spain. Association for Computational Linguistics.

\bibitem[{Lopez-Paz and Ranzato(2017)}]{lopez2017gradient}
David Lopez-Paz and Marc'Aurelio Ranzato. 2017.
\newblock Gradient episodic memory for continual learning.
\newblock \emph{Advances in neural information processing systems}, 30.

\bibitem[{Luo et~al.(2023)Luo, Yang, Meng, Li, Zhou, and Zhang}]{luo2023empirical}
Yun Luo, Zhen Yang, Fandong Meng, Yafu Li, Jie Zhou, and Yue Zhang. 2023.
\newblock \href {http://arxiv.org/abs/2308.08747} {An empirical study of catastrophic forgetting in large language models during continual fine-tuning}.

\bibitem[{Miao et~al.(2023)Miao, Zhang, Su, Li, Luan, Chen, Wang, and Zhang}]{miao-etal-2023-exploring}
Zhongjian Miao, Wen Zhang, Jinsong Su, Xiang Li, Jian Luan, Yidong Chen, Bin Wang, and Min Zhang. 2023.
\newblock \href {https://doi.org/10.18653/v1/2023.emnlp-main.178} {Exploring all-in-one knowledge distillation framework for neural machine translation}.
\newblock In \emph{Proceedings of the 2023 Conference on Empirical Methods in Natural Language Processing}, pages 2929--2940, Singapore. Association for Computational Linguistics.

\bibitem[{Mok et~al.(2023)Mok, Do, Lee, Taghavi, Yu, and Yoon}]{mok-etal-2023-large}
Jisoo Mok, Jaeyoung Do, Sungjin Lee, Tara Taghavi, Seunghak Yu, and Sungroh Yoon. 2023.
\newblock \href {https://doi.org/10.18653/v1/2023.acl-long.703} {Large-scale lifelong learning of in-context instructions and how to tackle it}.
\newblock In \emph{Proceedings of the 61st Annual Meeting of the Association for Computational Linguistics (Volume 1: Long Papers)}, pages 12573--12589, Toronto, Canada. Association for Computational Linguistics.

\bibitem[{OpenAI(2023)}]{openai2023gpt4}
OpenAI. 2023.
\newblock \href {http://arxiv.org/abs/2303.08774} {Gpt-4 technical report}.

\bibitem[{Razdaibiedina et~al.(2023)Razdaibiedina, Mao, Hou, Khabsa, Lewis, and Almahairi}]{razdaibiedina2023progressive}
Anastasia Razdaibiedina, Yuning Mao, Rui Hou, Madian Khabsa, Mike Lewis, and Amjad Almahairi. 2023.
\newblock \href {http://arxiv.org/abs/2301.12314} {Progressive prompts: Continual learning for language models}.

\bibitem[{Rolnick et~al.(2019)Rolnick, Ahuja, Schwarz, Lillicrap, and Wayne}]{rolnick2019experience}
David Rolnick, Arun Ahuja, Jonathan Schwarz, Timothy~P. Lillicrap, and Greg Wayne. 2019.
\newblock \href {http://arxiv.org/abs/1811.11682} {Experience replay for continual learning}.

\bibitem[{Scialom et~al.(2022)Scialom, Chakrabarty, and Muresan}]{scialom-etal-2022-fine}
Thomas Scialom, Tuhin Chakrabarty, and Smaranda Muresan. 2022.
\newblock \href {https://doi.org/10.18653/v1/2022.emnlp-main.410} {Fine-tuned language models are continual learners}.
\newblock In \emph{Proceedings of the 2022 Conference on Empirical Methods in Natural Language Processing}, pages 6107--6122, Abu Dhabi, United Arab Emirates. Association for Computational Linguistics.

\bibitem[{Smith et~al.(2021)Smith, Hsu, Balloch, Shen, Jin, and Kira}]{smith2021dreaming}
James Smith, Yen-Chang Hsu, Jonathan Balloch, Yilin Shen, Hongxia Jin, and Zsolt Kira. 2021.
\newblock \href {http://arxiv.org/abs/2106.09701} {Always be dreaming: A new approach for data-free class-incremental learning}.

\bibitem[{Taori et~al.(2023)Taori, Gulrajani, Zhang, Dubois, Li, Guestrin, Liang, and Hashimoto}]{alpaca}
Rohan Taori, Ishaan Gulrajani, Tianyi Zhang, Yann Dubois, Xuechen Li, Carlos Guestrin, Percy Liang, and Tatsunori~B. Hashimoto. 2023.
\newblock Stanford alpaca: An instruction-following llama model.
\newblock \url{https://github.com/tatsu-lab/stanford_alpaca}.

\bibitem[{Touvron et~al.(2023{\natexlab{a}})Touvron, Lavril, Izacard, Martinet, Lachaux, Lacroix, Rozière, Goyal, Hambro, Azhar, Rodriguez, Joulin, Grave, and Lample}]{touvron2023llama}
Hugo Touvron, Thibaut Lavril, Gautier Izacard, Xavier Martinet, Marie-Anne Lachaux, Timothée Lacroix, Baptiste Rozière, Naman Goyal, Eric Hambro, Faisal Azhar, Aurelien Rodriguez, Armand Joulin, Edouard Grave, and Guillaume Lample. 2023{\natexlab{a}}.
\newblock \href {http://arxiv.org/abs/2302.13971} {Llama: Open and efficient foundation language models}.

\bibitem[{Touvron et~al.(2023{\natexlab{b}})Touvron, Martin, Stone, Albert, Almahairi, Babaei, Bashlykov, Batra, Bhargava, Bhosale, Bikel, Blecher, Ferrer, Chen, Cucurull, Esiobu, Fernandes, Fu, Fu, Fuller, Gao, Goswami, Goyal, Hartshorn, Hosseini, Hou, Inan, Kardas, Kerkez, Khabsa, Kloumann, Korenev, Koura, Lachaux, Lavril, Lee, Liskovich, Lu, Mao, Martinet, Mihaylov, Mishra, Molybog, Nie, Poulton, Reizenstein, Rungta, Saladi, Schelten, Silva, Smith, Subramanian, Tan, Tang, Taylor, Williams, Kuan, Xu, Yan, Zarov, Zhang, Fan, Kambadur, Narang, Rodriguez, Stojnic, Edunov, and Scialom}]{touvron2023llama2}
Hugo Touvron, Louis Martin, Kevin Stone, Peter Albert, Amjad Almahairi, Yasmine Babaei, Nikolay Bashlykov, Soumya Batra, Prajjwal Bhargava, Shruti Bhosale, Dan Bikel, Lukas Blecher, Cristian~Canton Ferrer, Moya Chen, Guillem Cucurull, David Esiobu, Jude Fernandes, Jeremy Fu, Wenyin Fu, Brian Fuller, Cynthia Gao, Vedanuj Goswami, Naman Goyal, Anthony Hartshorn, Saghar Hosseini, Rui Hou, Hakan Inan, Marcin Kardas, Viktor Kerkez, Madian Khabsa, Isabel Kloumann, Artem Korenev, Punit~Singh Koura, Marie-Anne Lachaux, Thibaut Lavril, Jenya Lee, Diana Liskovich, Yinghai Lu, Yuning Mao, Xavier Martinet, Todor Mihaylov, Pushkar Mishra, Igor Molybog, Yixin Nie, Andrew Poulton, Jeremy Reizenstein, Rashi Rungta, Kalyan Saladi, Alan Schelten, Ruan Silva, Eric~Michael Smith, Ranjan Subramanian, Xiaoqing~Ellen Tan, Binh Tang, Ross Taylor, Adina Williams, Jian~Xiang Kuan, Puxin Xu, Zheng Yan, Iliyan Zarov, Yuchen Zhang, Angela Fan, Melanie Kambadur, Sharan Narang, Aurelien Rodriguez, Robert Stojnic, Sergey Edunov, and Thomas
  Scialom. 2023{\natexlab{b}}.
\newblock \href {http://arxiv.org/abs/2307.09288} {Llama 2: Open foundation and fine-tuned chat models}.

\bibitem[{Wang et~al.(2023)Wang, Si, Li, Lukasik, Yu, Hsieh, Dhillon, and Kumar}]{wang2023twostage}
Yihan Wang, Si~Si, Daliang Li, Michal Lukasik, Felix Yu, Cho-Jui Hsieh, Inderjit~S Dhillon, and Sanjiv Kumar. 2023.
\newblock \href {http://arxiv.org/abs/2211.00635} {Two-stage llm fine-tuning with less specialization and more generalization}.

\bibitem[{Wang et~al.(2022)Wang, Mishra, Alipoormolabashi, Kordi, Mirzaei, Naik, Ashok, Dhanasekaran, Arunkumar, Stap, Pathak, Karamanolakis, Lai, Purohit, Mondal, Anderson, Kuznia, Doshi, Pal, Patel, Moradshahi, Parmar, Purohit, Varshney, Kaza, Verma, Puri, Karia, Doshi, Sampat, Mishra, Reddy~A, Patro, Dixit, and Shen}]{wang-etal-2022-super}
Yizhong Wang, Swaroop Mishra, Pegah Alipoormolabashi, Yeganeh Kordi, Amirreza Mirzaei, Atharva Naik, Arjun Ashok, Arut~Selvan Dhanasekaran, Anjana Arunkumar, David Stap, Eshaan Pathak, Giannis Karamanolakis, Haizhi Lai, Ishan Purohit, Ishani Mondal, Jacob Anderson, Kirby Kuznia, Krima Doshi, Kuntal~Kumar Pal, Maitreya Patel, Mehrad Moradshahi, Mihir Parmar, Mirali Purohit, Neeraj Varshney, Phani~Rohitha Kaza, Pulkit Verma, Ravsehaj~Singh Puri, Rushang Karia, Savan Doshi, Shailaja~Keyur Sampat, Siddhartha Mishra, Sujan Reddy~A, Sumanta Patro, Tanay Dixit, and Xudong Shen. 2022.
\newblock \href {https://doi.org/10.18653/v1/2022.emnlp-main.340} {Super-{N}atural{I}nstructions: Generalization via declarative instructions on 1600+ {NLP} tasks}.
\newblock In \emph{Proceedings of the 2022 Conference on Empirical Methods in Natural Language Processing}, pages 5085--5109, Abu Dhabi, United Arab Emirates. Association for Computational Linguistics.

\bibitem[{Xu and Zhu(2018)}]{xu2018reinforced}
Ju~Xu and Zhanxing Zhu. 2018.
\newblock \href {http://arxiv.org/abs/1805.12369} {Reinforced continual learning}.

\bibitem[{Yin et~al.(2020)Yin, Molchanov, Li, Alvarez, Mallya, Hoiem, Jha, and Kautz}]{yin2020dreaming}
Hongxu Yin, Pavlo Molchanov, Zhizhong Li, Jose~M. Alvarez, Arun Mallya, Derek Hoiem, Niraj~K. Jha, and Jan Kautz. 2020.
\newblock \href {http://arxiv.org/abs/1912.08795} {Dreaming to distill: Data-free knowledge transfer via deepinversion}.

\bibitem[{Zhang et~al.(2022)Zhang, Zhang, Xiang, Liang, Su, Miao, Wang, and Xu}]{zhang-etal-2022-clle}
Han Zhang, Sheng Zhang, Yang Xiang, Bin Liang, Jinsong Su, Zhongjian Miao, Hui Wang, and Ruifeng Xu. 2022.
\newblock \href {https://doi.org/10.18653/v1/2022.findings-emnlp.30} {{CLLE}: A benchmark for continual language learning evaluation in multilingual machine translation}.
\newblock In \emph{Findings of the Association for Computational Linguistics: EMNLP 2022}, pages 428--443, Abu Dhabi, United Arab Emirates. Association for Computational Linguistics.

\bibitem[{Zhang et~al.(2023{\natexlab{a}})Zhang, Su, Min, Miao, Hu, Fu, Shi, and Chen}]{zhang2023exploring}
Liang Zhang, Jinsong Su, Zijun Min, Zhongjian Miao, Qingguo Hu, Biao Fu, Xiaodong Shi, and Yidong Chen. 2023{\natexlab{a}}.
\newblock \href {https://doi.org/10.1609/aaai.v37i11.26635} {Exploring self-distillation based relational reasoning training for document-level relation extraction}.
\newblock In \emph{Proceedings of the AAAI Conference on Artificial Intelligence}, volume~37, pages 13967--13975.

\bibitem[{Zhang et~al.(2023{\natexlab{b}})Zhang, Fang, Chen, and Namazi-Rad}]{zhang-etal-2023-citb}
Zihan Zhang, Meng Fang, Ling Chen, and Mohammad-Reza Namazi-Rad. 2023{\natexlab{b}}.
\newblock \href {https://doi.org/10.18653/v1/2023.findings-emnlp.633} {{CITB}: A benchmark for continual instruction tuning}.
\newblock In \emph{Findings of the Association for Computational Linguistics: EMNLP 2023}, pages 9443--9455, Singapore. Association for Computational Linguistics.

\end{thebibliography}
\bibliographystyle{acl_natbib}

\clearpage

\appendix

\section{Details of the Selected 10 SuperNI Tasks}
\label{app:task-info}
Table \ref{tab:task-info} lists all of the selected 10 SuperNI tasks for our main experiments. To simplify the description, we utilize abbreviations to represent these tasks in this paper.
The SuperNI dataset can be found at \url{https://github.com/allenai/natural-instructions}.

\section{More Details of Experiments on 5 SuperNI Tasks}
\label{app:five-task}

Table \ref{tab:cl-order} lists all of the continual learning orders on 5 SuperNI tasks conducted in our experiments.

\begin{table}[htbp]
    \tabcolsep=7pt
    \small
    \centering
    \begin{tabular}{cp{0.5cm}c}
        \toprule
        \textbf{Order} & & \textbf{Task Sequence} \\
        \midrule
         1 & & QA $\rightarrow$ QG $\rightarrow$ SA $\rightarrow$ Sum. $\rightarrow$ Trans. \\
         2 & & Trans. $\rightarrow$ SA $\rightarrow$ QA $\rightarrow$ Sum. $\rightarrow$ QG \\
         3 & & Sum. $\rightarrow$ QG $\rightarrow$ Trans. $\rightarrow$ QA $\rightarrow$ SA \\
         \bottomrule
    \end{tabular}
    \caption{Continual learning orders on 5 SuperNI tasks.}
    \label{tab:cl-order}
\end{table}

Table \ref{tab:five-task-fwt} shows the final FWT results on 5 SuperNI tasks. Our SSR framework surpasses $r=$~1\% rehearsal-based baselines but falls behind $r=$~10\% counterparts. However, as illustrated in Figure \ref{fig:overall}, the FWT performance of SSR will finally surpass the $r=$~10\% rehearsal-based baselines as the number of training stages increases.

\begin{table}[htbp]
\tabcolsep=4pt
\small
\centering
\begin{tabular}{lcccccccccccc}
\toprule
\textbf{Model}
& \textbf{Order 1} & \textbf{Order 2} & \textbf{Order 3} & \textbf{Avg.} \\\midrule
\multicolumn{5}{c}{\textit{Llama-2-7b}} \\
\midrule
Non-rehearsal & 2.49 & 1.99 & 3.42 & 2.63 \\
\hdashline
RandSel(1\%) & 3.77 & 2.84 & 4.21 & 3.61\\
KMeansSel(1\%) & 4.75 & 2.46 & 5.25 & 4.15 \\
RandSel(10\%) & \textbf{10.34} & \textbf{3.20} & \textbf{7.54} & \textbf{7.03} \\
KMeansSel(10\%) & 7.61 & 2.67 & 6.50 & 5.59 \\
\hdashline
SSR & 7.14 & 2.62 & 6.39 & 5.38 \\
\midrule
\multicolumn{5}{c}{\textit{Llama-2-7b-chat}} \\
\midrule
Non-rehearsal & 16.93 & 13.56 & 22.13 & 17.54 \\
\hdashline
RandSel(1\%) & 19.06 & 17.14 & 18.67 & 18.29 \\
KMeansSel(1\%) & 23.19 & 16.46 & 18.36 & 19.34 \\
RandSel(10\%) & 26.29 & \textbf{17.52} & 19.53 & 21.11 \\
KMeansSel(10\%) & \textbf{27.05} & 16.65 & 18.59 & 20.76 \\
\hdashline
SSR & 26.63 & 16.41 & \textbf{26.10} & \textbf{23.05} \\
\midrule
\multicolumn{5}{c}{\textit{Alpaca-7b}} \\
\midrule
Non-rehearsal & 22.32 & 13.13 & 21.88 & 19.11 \\
\hdashline
RandSel(1\%) & 25.26 & 14.62 & 25.46 & 21.78 \\
KMeansSel(1\%) & 25.07 & 15.10 & 25.50 & 21.89 \\
RandSel(10\%) & 25.97 & 16.76 & \textbf{30.34} & \textbf{24.36} \\
KMeansSel(10\%) & \textbf{26.41} & 17.14 & 27.01 & 23.52 \\
\hdashline
SSR & 25.45 & \textbf{17.45} & 27.69 & 23.53 \\
\bottomrule
\end{tabular}
\caption{Final FWT results on 5 SuperNI tasks under different continual learning orders.}
\label{tab:five-task-fwt}
\end{table}

\section{More Details of Experiments on 10 SuperNI Tasks}
\label{app:ten-task}

\begin{figure*}[t]
    \centering \includegraphics[width=\linewidth]{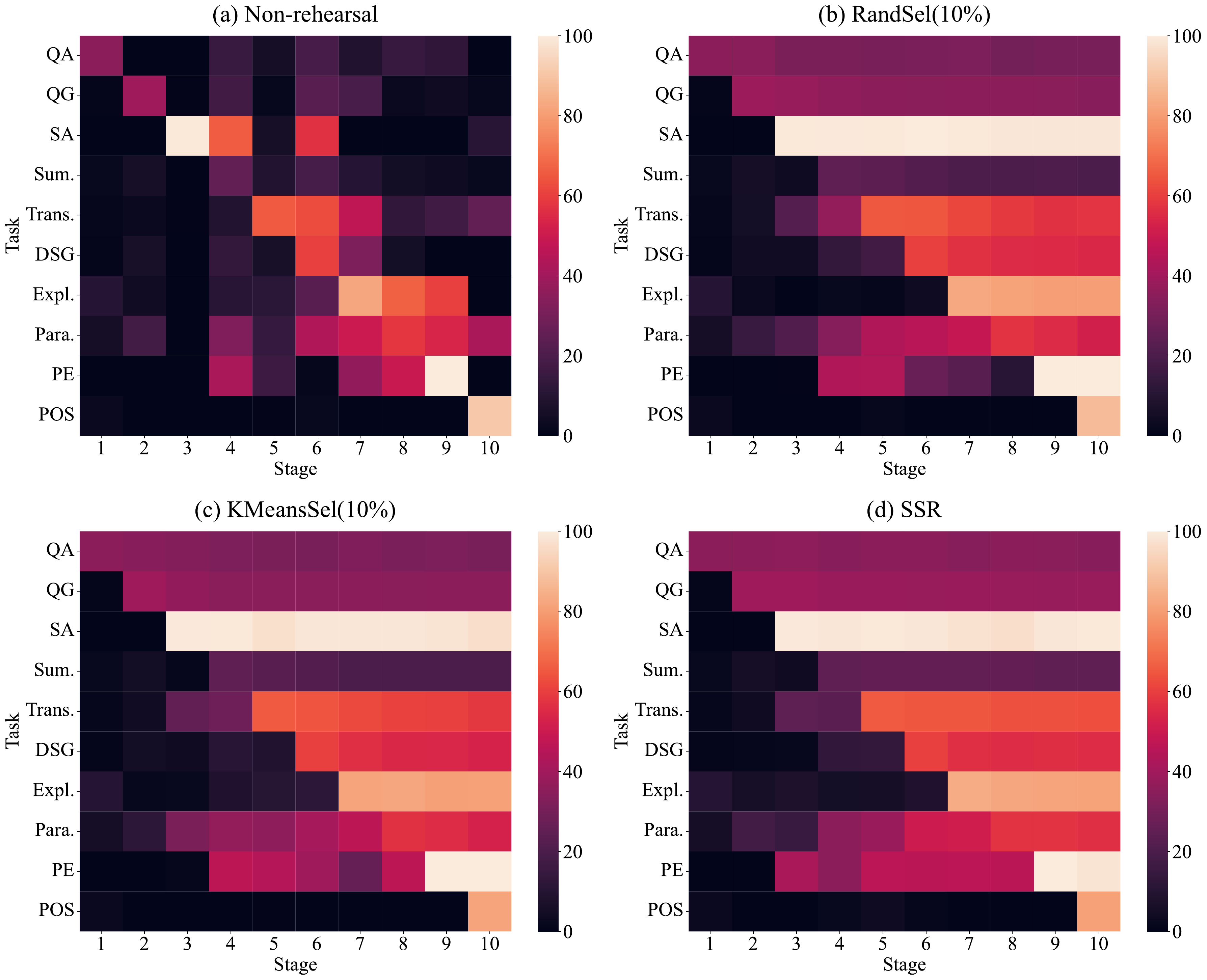}
    \caption{ROUGE-L heatmaps for Llama-2-7b on 10 SuperNI tasks.}
    \label{fig:heatmap}
\end{figure*}

Figure \ref{fig:heatmap} depicts the heatmaps of the ROUGE-L performance for Llama-2-7b across 10 SuperNI tasks. A visual inspection reveals that the non-rehearsal baseline rapidly forgets previously learned tasks when subsequent tasks are learned. In contrast, after learning a task in its respective stage, SSR demonstrates minimal color change in future stages, indicating the least amount of forgetting.

Table \ref{tab:ten-task-other} highlights that SSR retains its superiority in terms of AR and BWT for Llama-2-7b-chat and Alpaca-7b on 10 SuperNI tasks. However, its FWT performance is comparable or inferior to that of rehearsal-based baselines. It is worth noting that Alpaca-7b tends to exhibit poorer performance regardless of the CL approaches employed.

\begin{table}[h]
\centering
\small
\tabcolsep=8pt
\begin{tabular}{lccc}
\toprule
\textbf{Model}& \textbf{AR} & \textbf{FWT} & \textbf{BWT} \\
\midrule
\multicolumn{4}{c}{\textit{Llama-2-7b-chat}} \\\midrule
MTL
& 62.68 & - & - \\
Non-rehearsal 
& 47.05 & 22.24 & -20.73 \\
\hdashline
RandSel(1\%)
& 60.70 & 25.77 & -4.17 \\
KMeansSel(1\%) 
& 61.53 & 26.62 & -4.24 \\
RandSel(10\%)
& 58.46 & \textbf{27.99} & -3.34 \\
KMeansSel(10\%) & 59.27 & 27.52 & -3.12 \\
\hdashline
SSR
& \textbf{63.34} & 27.70 & \textbf{-1.27} \\
\midrule
\multicolumn{4}{c}{\textit{Alpaca-7b}} \\\midrule
MTL
& 63.60 & - & - \\
Non-rehearsal 
& 36.37 & 25.44 & -31.50 \\
\hdashline
RandSel(1\%)
& 59.99 & \textbf{28.95} & -4.58 \\
KMeansSel(1\%) 
& 58.70 & 27.53 & -3.53 \\
RandSel(10\%)
& 58.93 & 28.11 & -3.16 \\
KMeansSel(10\%) & 58.72 & 27.93 & -3.09 \\
\hdashline
SSR
& \textbf{60.24} & 27.55 & \textbf{-2.35} \\
\bottomrule
\end{tabular}
\caption{Final results for Llama-2-7b-chat and Alpaca-7b on 10 SuperNI tasks.}
\label{tab:ten-task-other}
\end{table}

\section{More Implementation Details of Experiments on the Generalization Capability Preservation of Alpaca-7b}
\label{app:alpaca-exp}

The continual learning order is as follows: (Alpaca-52k $\rightarrow$) QA $\rightarrow$ QG $\rightarrow$ SA $\rightarrow$ Sum. $\rightarrow$ Trans. To conduct ICL, we utilize 1\% / 0.1\% of the training data from SuperNI tasks and Alpaca-52k as demonstrations, respectively. When clustering instances, we use KMeans with $C=$ 20 clusters for synthetic instances of SuperNI tasks and $C=$ 520 for those of Alpaca-52k. We retain synthetic instances with the SSR ratio $\hat{r}=$ 10\% / 1\% for SuperNI tasks and Alpaca-52k, respectively. 

\begin{table*}[t]
\centering
\small
\begin{tabular}{lccc}
\toprule
\textbf{Abbr.} &
\textbf{Category} &
\textbf{Name} &
\textbf{NLU task} \\
\midrule
QA & Question Answering & task024\_cosmosqa\_answer\_generation & - \\
QG & Question Generation & task074\_squad1.1\_question\_generation & - \\
SA & Sentiment Analysis & task1312\_amazonreview\_polarity\_classification & + \\
Sum. & Summarization & task511\_reddit\_tifu\_long\_text\_summarization & - \\
Trans. & Translation & task1219\_ted\_translation\_en\_es & - \\
DSG & Dialogue Sentence Generation & task574\_air\_dialogue\_sentence\_generation & - \\
Expl. & Explanation & task192\_hotpotqa\_sentence\_generation & - \\
Para. & Paraphrasing & task177\_para-nmt\_paraphrasing & - \\
POS & POS Tagging & task346\_hybridqa\_classification & + \\
PE & Program Execution & task064\_all\_elements\_except\_first\_i & - \\
\bottomrule
\end{tabular}
\caption{Details of the selected 10 SuperNI tasks.}
\label{tab:task-info} 
\end{table*}

\end{document}